# Sparse Concept Coded Tetrolet Transform for Unconstrained Odia Character Recognition

Kalyan S Dash, N B Puhan, G Panda

School of Electrical Sciences,
Indian Institute of Technology Bhubaneswar
Odisha - 751013, India.
{ksd10, nbpuhan, gpanda}@iitbbs.ac.in

*Abstract*— **Feature representation in the form of spatio-spectral decomposition is one of the robust techniques adopted in automatic handwritten character recognition systems. In this regard, we propose a new image representation approach for unconstrained handwritten alphanumeric characters using sparse concept coded Tetrolets. Tetrolets, which does not use fixed dyadic square blocks for spectral decomposition like conventional wavelets, preserve the localized variations in handwritings by adopting tetrominoes those capture the shape geometry. The sparse concept coding of low entropy Tetrolet representation is found to extract the important hidden information (concept) for superior pattern discrimination. Large scale experimentation using ten databases in six different scripts (Bangla, Devanagari, Odia, English, Arabic and Telugu) has been performed. The proposed feature representation along with standard classifiers such as random forest, support vector machine (SVM), nearest neighbor and modified quadratic discriminant function (MQDF) is found to achieve state-of-the-art recognition performance in all the databases, viz. 99.40% (MNIST); 98.72% and 93.24% (IITBBS); 99.38% and 99.22% (ISI Kolkata). The proposed OCR system is shown to perform better than other sparse based techniques such as PCA, SparsePCA and SparseLDA, as well as better than existing transforms (Wavelet, Slantlet and Stockwell).**

# Sparse Concept Coded Tetrolet Transform for Unconstrained Odia Character Recognition

Kalyan S. Dash, N. B. Puhan, and Ganapati Panda

*Abstract*— **Feature representation in the form of spatio-spectral decomposition is one of the robust techniques adopted in automatic handwritten character recognition systems. In this regard, we propose a new image representation approach for unconstrained handwritten alphanumeric characters using sparse concept coded Tetrolets. Tetrolets, which does not use fixed dyadic square blocks for spectral decomposition like conventional wavelets, preserve the localized variations in handwritings by adopting tetrominoes those capture the shape geometry. The sparse concept coding of low entropy Tetrolet representation is found to extract the important hidden information (concept) for superior pattern discrimination. Large scale experimentation using ten databases in six different scripts (Bangla, Devanagari, Odia, English, Arabic and Telugu) has been performed. The proposed feature representation along with standard classifiers such as random forest, support vector machine (SVM), nearest neighbor and modified quadratic discriminant function (MQDF) is found to achieve state-of-the-art recognition performance in all the databases, viz. 99.40% (MNIST); 98.72% and 93.24% (IITBBS); 99.38% and 99.22% (ISI Kolkata). The proposed OCR system is shown to perform better than other sparse based techniques such as PCA, SparsePCA and SparseLDA, as well as better than existing transforms (Wavelet, Slantlet and Stockwell).**

## 1. Introduction

MACHINES those can mimic human reading and comprehension is far from reality, particularly in case of unconstrained handwritings. This is because of varying writing styles of individuals which introduce nonlinear deformations in handwriting patterns. A conventional character recognition system comprises basically of three blocks: pre-processor, feature extractor and recognizer or classifier. Among these, the feature extractor block is considered to be of significant importance as feature representation is one of the most influential aspects of pattern recognition. Especially so, when there is large variation in intraclass patterns and presence of perceptually similar interclass patterns as found in handwritten characters.

State-of-the-art handwritten OCR systems represent images either directly in the spatial domain [1-2], by extracting features from shape geometry [3-5], or using several image transforms [6-9]. Popularly used spatial domain features include curvature [10], quad-tree based longest run (QTLR) [11], gradient and directional features [12-14], centroids [15], moments [16], morphological [17] and features learnt from raw image in deep learning framework [18]. As frequency content is the most commonly searched attribute of an image, a number of frequency domain transformations is also available in the literatures of OCR systems [19].

Research contributions relating to Odia handwritten OCR include feature based training and recognition approach as proposed by Pal et al. [10, 20], Bhowmik et al. [21], Roy et al. [22], Dash et al. [14, 23], etc. on one hand and Gestalt configural superiority effect that does not require exhaustive apriori training [24] on the other. These methods use various features such as gradient and directional (chaincode) features [14, 20, 22], scalar features [21], and transform domain features [23, 25]. In contrast, Bangla OCR has got more attention and has even contributed to partial automation of postal services in Bangladesh. Researchers have tried to explore many feature extraction techniques including directional, structural, statistical, topological, morphological, transform domain based [11-13, 16-17, 26-27]. The classifiers used include binary tree, neural networks, support vector machine, nearest neighbor, several versions of quadratic discriminant function classifier (MQDF, DLQDF).

Existing state-of-the-art work on MNIST database reports various feature extraction techniques such as active graphs [28], shape context [29], directional features [30], and features trained from raw image pixels using deep convolutional neural networks [18]. Classifiers such as nearest neighbor, support vector machine (SVM), neural networks have shown to achieve high accuracies on English handwritten numerals [18, 28-30]. Similarly, existing literature on recognition of other Asian languages such as Telugu, Arabic and Devanagari follow a similar approach of feature extraction and recognition models that can be found in [15, 31].

The basic structure of characters are represented by low frequency coefficients whereas the high frequencies hold the detailed variations. Because of the global nature of Fourier transform which includes frequencies over the whole spectrum, several multiresolution spatial and spectral representations are developed which capture the localized deformations in handwritings. Such space-frequency decomposition techniques prove to effectively represent character patterns and several research works have been reported using Gabor transform [32], Contourlets [33], Wavelets [26], Slantlets [25], Stockwell transform [23].

The issue with existing spectral domain or spatio-spectral decomposition based feature extraction techniques such as wavelets is that of their fixed block dyadic structures. Such fixed blocking squares are proved to be inefficient because they disregard the local structures in an image. This localized information is very important in case of unconstrained handwriting analysis. The motivation behind the space-frequency decomposition using Tetrolets [34] for character recognition is to adopt a more general partitioning such that local image geometry is taken into account while extracting the spectral information. Tetrolets have been shown to achieve superior performance in other applications such as image fusion [35] and denoising [36].

The representation is known to be robust if it maximizes interclass pattern separation as well as minimises intraclass pattern variability. At the same time the feature space should not suffer from the curse of dimensionality which hampers the recognition speed and demands more memory storage. PCA and other similar feature reduction based techniques in character recognition have been proposed in [37-38]. However, there is a trade off as to how small the feature dimension can be and how discriminative it should be for achieving the desired goal of high accuracy. To address both these concerns, we propose a sparse based space-frequency (Tetrolet) representation of handwritten characters and numerals. Our contribution in this paper is three-fold:

1) A low entropy non-redundant Tetrolet based spatio-spectral decomposition is adopted which is sensitive to local shape geometry of handwritten images. This multiresolution decomposition uses tetrominoes [39] as fundamental basis to represent the characters in a spatio-spectral manner.
2) Sparse concept coding (SCC) [40] is applied to the transform coefficients to extract the hidden 'concepts' or the most discriminative features while satisfying dimensionality reduction.
3) Extensive performance evaluation is carried out in ten databases comprising of English, Arabic, Bangla, Devanagari, Odia and Telugu scripts. The proposed handwritten OCR system is found to achieve state-of-the-art classification results with high recognition speed in all databases.

The remainder of the paper is organized as follows: the proposed recognition system based on a sparse concept coded Tetrolet feature representation is described in Section II. The implementation setup, validation techniques and experimental results along with comparison studies are outlined in Section III. Section IV concludes the paper.

## 2. Proposed Sparse concept coded Tetrolet Representation

### *2.1. Tetrominoes*

Tetrominoes are geometric shapes made by connecting four equal-sized squares, each joined together with at least one other square along an edge and not merely at their corners. Originally introduced by Golomb [39], tetrominoes became popular in the famous computer game 'Tetris' [41]. There are five basic free tetrominoes disregarding rotations and reflections (Fig. 1).

For an image $\mathbf{a} = (a[i,j])_{(i,j)\in I}$ where $I = \{(i,j) : i, j = 0,...,N-1\} \subset \mathbf{Z}^2$ is the row-column index set with $N = 2^J$, $J \in \mathbf{N}$, a four neighbourhood of an index $(i,j) \in I$ is defined as:

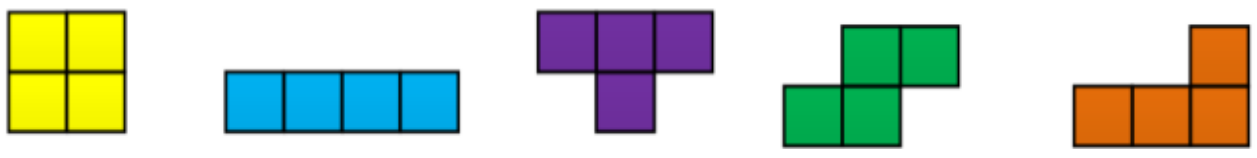

**Fig. 1**. Five basic free tetrominoes

$$N_4(i,j) := \{(i-1,j),(i+1,j),(i,j-1),(i,j+1)\}. \tag{1}$$

A bijective mapping $J$ is applied for one-dimensional indexing as $J: I \to \{0,1,\dots,N^2-1\}$ with $J((i,j)) := jN+i$. A disjoint partition set $\{I_0,\dots,I_\mu\}, \mu \in \mathbf{N}$ of subset $I_\alpha \subset I$ is defined if

$$I_\alpha \cap I_\beta = \phi \quad for \;\; \alpha \neq \beta,$$

$$\bigcup_{\alpha=0}^{\mu} I_\alpha = I \tag{2}$$

provided,

$$|I_\alpha| = 4,$$

$$\forall (i,j) \in I_\alpha \;\; \exists (i',j') \in I_\alpha : (i',j') \in N_4(i,j). \tag{3}$$

Such subsets $I_\alpha$ are called tetrominoes.

## 2.2. *From tetrominoes to Tetrolets*

Tetrolets are adaptive Haar wavelets with tetrominoes as support. Starting with our original input handwritten character image $\mathbf{a}^0 = (a[i,j])_{i,j=0}^{N-1}$ with $N = 2^J$, $J \in \mathbf{N}$, successive $J-1$ number of Tetrolets can be applied to obtain a spatio-spectral decomposition. At $m^{th}$ level, the low-pass image $\mathbf{a}^{m-1}$ is divided into blocks $Z_{i,j}$ of $4\times4$ size where $i,j = 0,\dots,\frac{N}{4^m}-1$. It is established that every square block $[0,N)^2$ can be covered by tetrominoes if and only if $N$ is even. For a $4\times4$ block, there are total 117 admissible solutions for disjoint covering with four tetrominoes [42].

For every admissible tetromino configuration $c = 1,\dots,117$ (Fig. 2), the low-pass and three high-pass parts are given as

$$\mathbf{a}^{m,(c)} = (a^{m,(c)}[\alpha])_{\alpha=0,1,2,3}$$

$$with \;\; a^{m,(c)}[\alpha] = \sum_{(x,y)\in I_\alpha^{(c)}} \varepsilon[0,\Gamma(x,y)]\, a^{m-1}[x,y] \tag{4}$$

$$\mathbf{w}_l^{m,(c)} = (w_l^{m,(c)}[\alpha])_{\substack{\alpha=0,1,2,3 \\ l=1,2,3}}$$

$$\text{with} \quad w_l^{m,(c)}[\alpha] = \sum_{(x,y)\in I_\alpha^{(c)}} \varepsilon\left[l,\Gamma(x,y)\right] a^{m-1}\left[x,y\right] \tag{5}$$

where the coefficients $\varepsilon\left[l,\Gamma(x,y)\right]$ are entries from the Haar wavelet transform matrix. The optimal covering $\left(c^*\right)$ that ensures minimal $l^1-norm$ is chosen such that for every block $Z_{i,j}$ we get an optimal Tetrolet decomposition $[a^{m,(c^*)}, w_1^{m,(c^*)}, w_2^{m,(c^*)}, w_3^{m,(c^*)}]$.

$$c^* = \arg\min_c \sum_{l=1}^{3} \left\|\mathbf{w}_l^{m,(c)}\right\|_1 = \arg\min_c \sum_{l=1}^{3}\sum_{\alpha=0}^{3} \left|w_l^{m,(c)}[\alpha]\right| \tag{6}$$

The orthonormal basis functions for decomposition of handwritten character images using Tetrolets are defined such that the resulting transformation is sparse, compact and non-redundant. For any tetromino $I_\alpha$, the basis functions are defined as

$$\Phi_{I_\alpha}\left[x,y\right] := \begin{cases} \frac{1}{2}, & (x,y)\in I_\alpha \\ 0, & otherwise \end{cases} \tag{7}$$

The character image block $\mathbf{a}\big|_Z = \left(a\left[x,y\right]\right)_{(x,y)\in Z}$ is then decomposed into detail and approximation parts (Fig. 3)

$$a\left[x,y\right] = \sum_{l=1}^{3}\sum_{\alpha=0}^{3}\sum_{(x',y')\in I_\alpha} \varepsilon\left[l,\Gamma\left(x',y'\right)\right] a\left[x',y'\right] \Psi_{I_\alpha}^l\left[x,y\right] + \frac{1}{2}\sum_{\alpha=0}^{3}\sum_{(x',y')\in I_\alpha} a\left[x',y'\right]\Phi_{I_\alpha}\left[x,y\right] \tag{8}$$

$$\Psi_{I_\alpha}^l\left[x,y\right] := \begin{cases} \varepsilon\left[l,\Gamma(x,y)\right], & (x,y)\in I_\alpha \\ 0, & otherwise \end{cases} \tag{9}$$

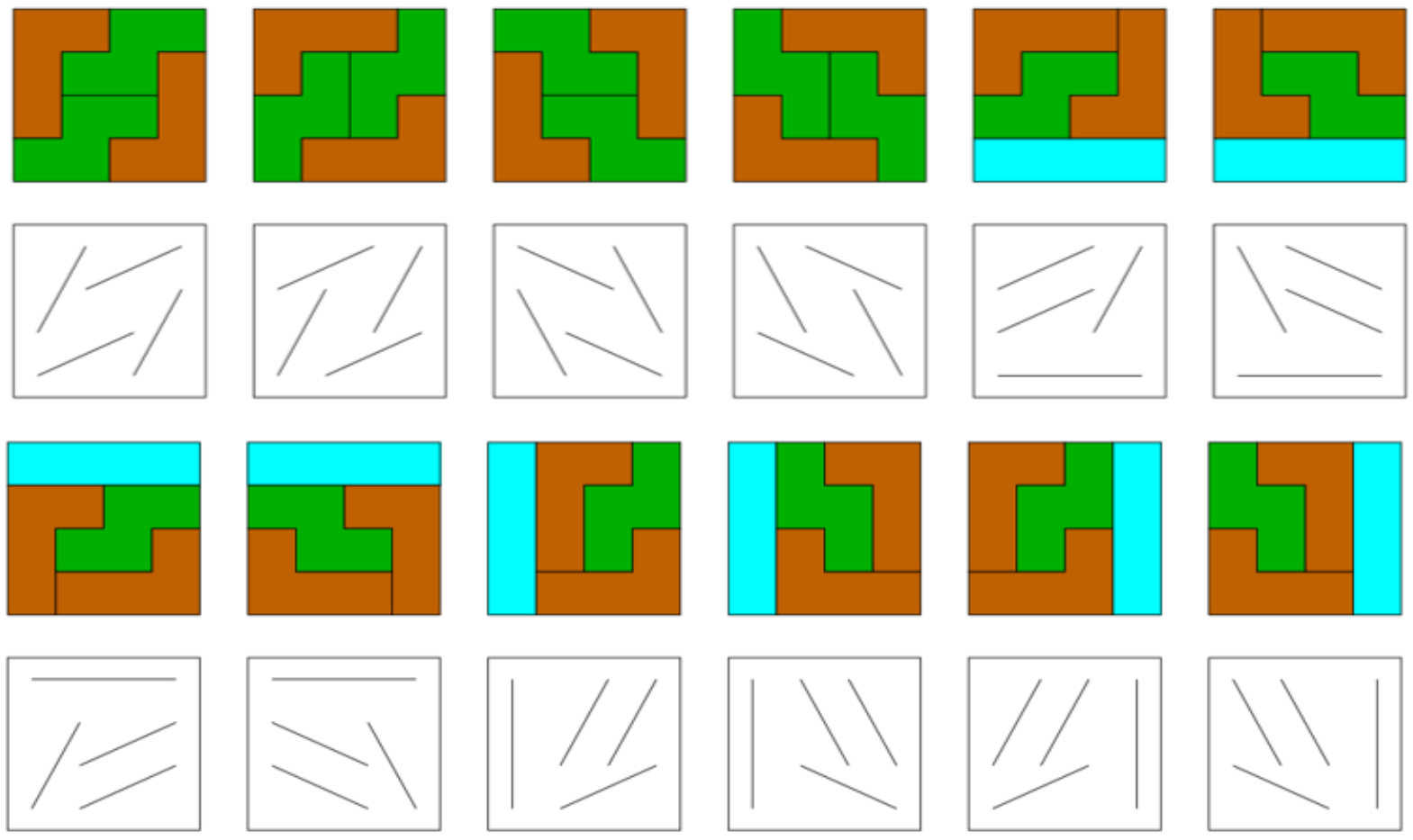

**Fig. 2.** Capturing different character stroke orientations using Tetrolet combinations

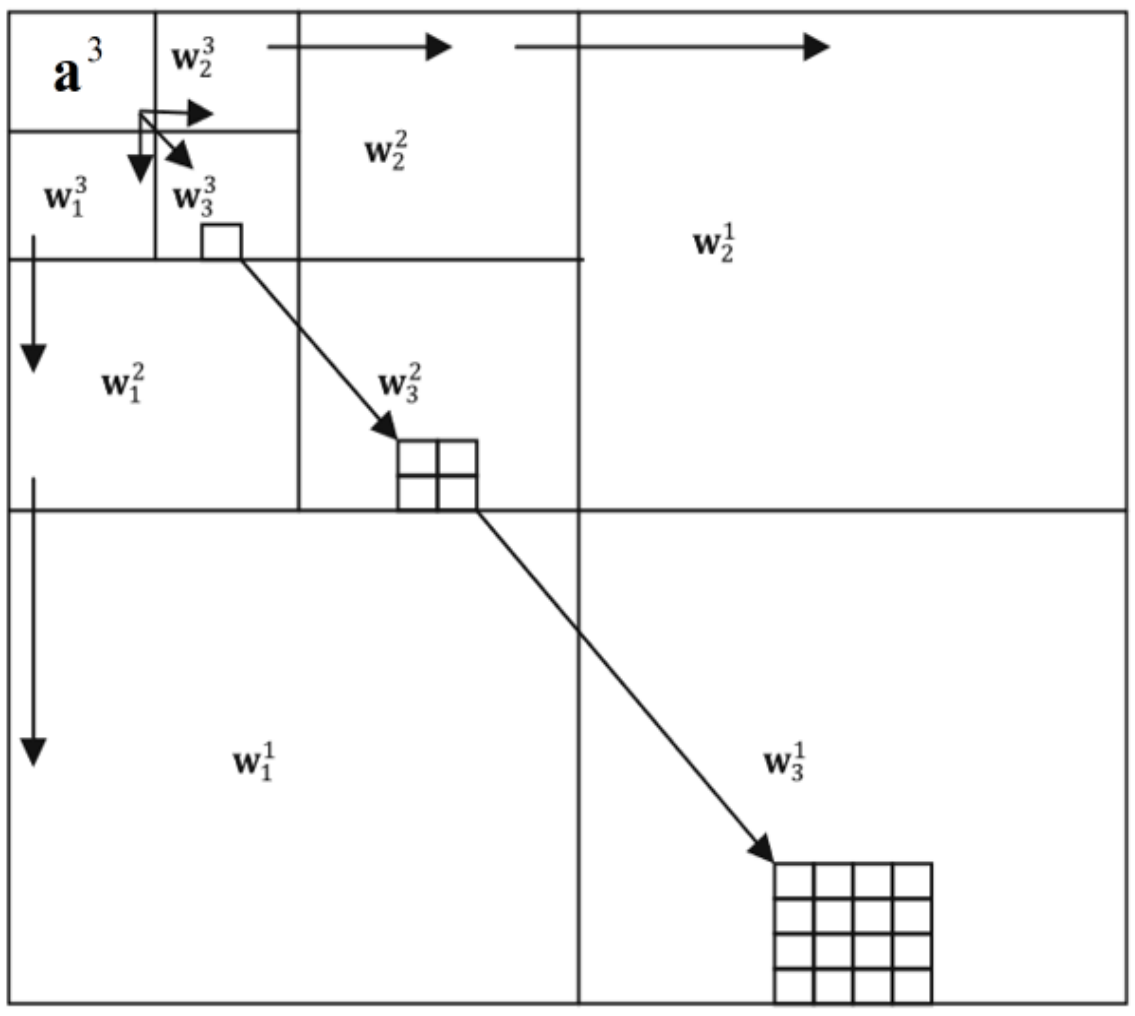

**Fig. 3.** Tetrolet subband regions

---

**Algorithm 1.** Image representation using Tetrolets

---

**Input**: $N \times N$ size normalised character image, $\left(N = 2^J\right)$

- Divide the image into $4 \times 4$ blocks
- Find the sparsest Tetrolet representation satisfying (13) in each block
- Rearrange the low- and high-pass coefficients of each block into a $2 \times 2$ block
- Store the high-pass coefficients $\mathbf{w}_l^{m,(c^*)}, l = 1,2,3$
- Apply step 1 to 4 to the low-pass image $\mathbf{a}^{m,(c^*)}$

**Output**: Tetrolet representation of character image

---

Finally a bivariate shrinkage function is applied on Tetrolet coefficients after the desired number of decomposition steps [43]. The function is defined as

$$(t)_+ = \begin{cases} t, & \text{if } t \geq 0 \\ 0, & \text{if } t < 0 \end{cases} \tag{10}$$

*2.3. Low entropy Tetrolet transform of character images*

The decomposition of an image into space-frequency transformation using Tetrolets (Fig. 4) reduces the number of wavelet coefficients as compared to the classical tensor product wavelets. But there is a trade off in terms of storing additional information. A low entropy model of Tetrolets is adopted to represent the handwritten character images by relaxing the optimal tetromino covering criterion. We know that higher the image dimension, more is the computation time and memory requirement for extracting robust features. A vector of length $N$ can be stored with $N \times B$ bits, where the bits per pixel, $(B)$ is defined as

$$B = -\sum_{\delta=1}^{\Delta} p(x_\delta)\log_2\left(p(x_\delta)\right). \tag{11}$$

For Tetrolets, there is a need to store $\frac{N^2}{4^{m+1}}$ covering values of tetromino configuration $c$ at the $m^{th}$ decomposition level. After $J = \log_2(N) - 1$ levels of decomposition, total additional storage requirement becomes

$$S = \frac{N^2}{12}\left(1 - \frac{1}{4^J}\right) = \frac{N^2 - 4}{12}. \tag{12}$$

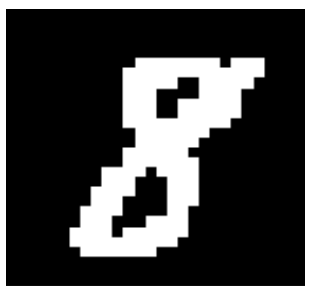

(a)

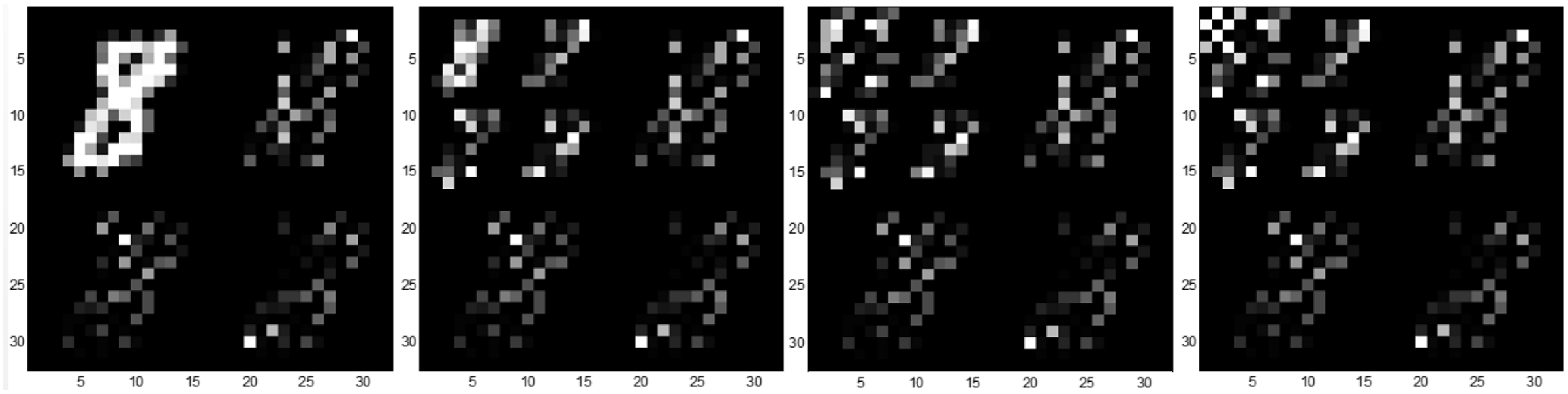

(b) (c) (d) (e)

**Fig. 4.** Spatio-spectral decomposition using Tetrolets: (a) original image from MNIST database, (b) level-1 decomposition, (c) level-2 decomposition, (d) level-3 decomposition, and (e) level-4 decomposition

We therefore allow a relaxation in the optimality condition of covering $(c^*)$ and find the set $C' \subset C$ of 'almost' optimal configurations $c$ that satisfy, with a tolerance threshold $\lambda$ .

$$\sum_{l=1}^{3}\sum_{\alpha=0}^{3}\left\|\mathbf{w}_l^{m,(c)}[\alpha]\right\|_1 \le \min_{c\in C}\sum_{l=1}^{3}\sum_{\alpha=0}^{3}\left\|\mathbf{w}_l^{m,(c)}[\alpha]\right\|_1 + \lambda \tag{13}$$

Among these configurations, the covering $c^* \in C'$ is picked which is chosen most frequently in the previous image block. Such representation achieves a satisfactory equilibrium between low adaptability cost and minimum Tetrolet coefficients.

### *2.4. Sparse concept coded feature extraction*

Once the input handwritten image is decomposed into a non-redundant space-frequency representation using Tetrolets, the transform domain coefficients can be fed to a recogniser as features. However, the size of such feature matrices poses a challenge to the performance of the classifier in terms of speed as well as accuracy. If the total number of alphanumeric character classes is $P$ with each class comprising $M$ number of $N \times N$ images, the feature matrix for training will be of $P \times M \times N^2$ dimension. We instead propose to generate a sparse coded feature matrix which not only reduces the computation overhead with faster training and recognition, but also ensures extraction of visually discriminative features (concepts) inspired by the considerable evidence that biological vision too adopts such sparse representation [44-45].

The most common approach for sparse coding has been matrix factorisation [46]. With $\mathbf{X}=[\mathbf{x}_1,...,\mathbf{x}_M] \in \mathbf{R}^{N^2 \times M}$ as the initial feature matrix for a certain character class, matrix factorisation generates two matrices $\mathbf{U} \in \mathbf{R}^{N^2 \times k}$ and $\mathbf{A} \in \mathbf{R}^{k \times M}$ such that

$$\mathbf{X} \approx \mathbf{UA} \quad \textit{satisfying} \quad \min_{\mathbf{U},\mathbf{A}} \|\mathbf{X} - \mathbf{UA}\|_2 \tag{14}$$

The columns of $\mathbf{U}$ forms the basis vectors which capture higher level features from the initial feature matrix whereas the columns of $\mathbf{A}$ are the $k$ – dimensional representation of original $N^2$-dimensional feature vector $(k << N^2)$. Rather than considering the Euclidean structure of the initial feature space for matrix factorisation, we obtain the set of best eigenvectors (basis) $\mathbf{Y}$ from sub-manifold learning of the ambient Euclidean space [47]. Then, the task becomes to learn the basis $\mathbf{U}$ which best fits $\mathbf{Y}$ by ridge regression:

$$\min_{\mathbf{U}} \|\mathbf{Y} - \mathbf{X}^T\mathbf{U}\|_2 + \tau\|\mathbf{U}\|_2 \tag{15}$$

where, $\tau$ is the regularisation parameter to avoid over-fitting.

The optimal solution for $\mathbf{U}$ is obtained by differentiating (15) with respect to $\mathbf{U}$ and equating to zero.

$$\mathbf{U}^* = \left(\mathbf{XX}^T + \tau\mathbf{I}\right)^{-1}\mathbf{XY} \tag{16}$$

Here $\mathbf{I}$ is the identity matrix. The sparse concept coded feature vectors $\mathbf{A}$ are now computed column by column using the $l_1$-regularised regression problem, *lasso* [35, 48] as

$$\min_{A_i} \left\| \mathbf{x}_i - \mathbf{U}^* A_i \right\|_2 + \rho \left| A_i \right| \tag{17}$$

where, $\mathbf{x}_i$ and $A_i$ are the $i^{th}$ columns of $\mathbf{X}$ and $\mathbf{A}$ respectively and $\rho$ controls the cardinality of the sparse feature vector.

### *2.5. Recognition model*

For each character class $p = 1,...,P$, a sparse concept coded feature matrix $\mathbf{A}$ is generated from a dictionary of $M$ training images. All such matrices form the training set $\mathbf{T} = \{\mathbf{A}_1, \mathbf{A}_2, ..., \mathbf{A}_P\}$. During recognition stage, an unknown (test) image is decomposed into its space-frequency transform using Tetrolets and is converted into $N^2 \times 1$ vector $\left(A^{tetrolet}\right)$. The high dimensional vector is projected onto the sparse feature subspaces using the learned basis $\mathbf{U}^*$ to obtain the sparse concept coded test feature vector $A^{scc}$.

$$A^{scc}{}_{k\times 1} = \left( \left(A^{tetrolet}\right)^T{}_{1\times N^2} \mathbf{U}^*{}_{N^2 \times k} \right)^T \tag{18}$$

An unknown test image is given a score $s$ based on the $l_2$-normalised difference between $A^{scc}$ and the columns of training set $\mathbf{T}$ during classification.

$$s\left(A^{scc}, \mathbf{A}_p\right) = \min_{A_j \in \mathbf{A}_p} \left\| \frac{A^{scc}}{\left\|A^{scc}\right\|_1} - \frac{A_j}{\left\|A_j\right\|_1} \right\|_2, \quad p = \{1,...,P\}, \; j = \{1,...,M\} \tag{19}$$

Finally, the label of the test character image is assigned as the character class which achieves the minimal relevance score (Fig. 5):

$$recognised_class\left(A^{scc}\right) = \arg\min_{p\in[1...P]} s\left(A^{scc}, \mathbf{A}_p\right) \tag{20}$$

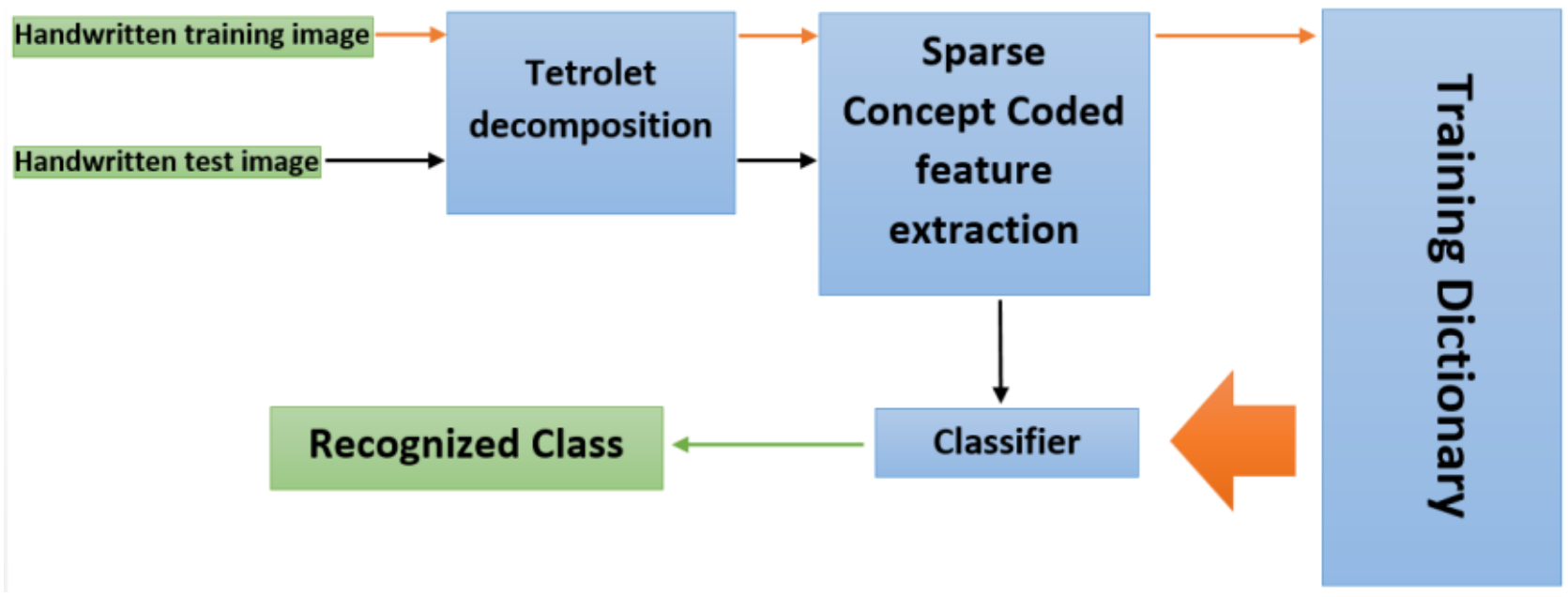

**Fig. 5**. Proposed handwritten character recognition system

## 3. RESULTS AND DISCUSSION

In this section, the proposed handwritten character recognition approach is validated against 10 benchmark databases of six different scripts such as Bangla, Devanagari, Odia, Telugu, Arabic and English. The implementation details are outlined below along with database description.

### *3.1. Implementation details*

The handwritten images are size normalized to $(32 \times 32)$ dimension to have uniformity as well as satisfying the $N = 2^n$ condition $(n = 5)$. The number of Tetrolet decomposition level is set at $J = \log_2(N) - 1 = 4$. The relaxation parameter $\lambda$ for finding low entropy optimal covering is assigned the value of 25 from experimentation. The sparse concept coded feature vector dimension is varied as $k = 64, 100, 200, 300, 400$ with different thresholds to select best eigenvalues corresponding to $\mathbf{Y}$. The size of the training and test data of the standard databases considered for experimental validation is given in Table I. The experiments are carried out on an Intel-*i7* processor with 8GB RAM in MATLAB environment.

### *3.2. Databases*

#### *3.2.1. CMATERdb3 database*

The CMATERdb3 database, developed by researchers at Jadavpur University, comprises of several handwritten alphanumeric characters from different languages and particularly those used in Indian subcontinent [31, 49]. The images are collected from three different sources, viz. class notes of students of different age groups, handwritten manuscripts of popular magazines, and from a pre-formatted data sheet specially designed for collection of handwriting samples. These documents are digitized using HP F380 flatbed scanner at 300 dpi. We consider five datasets of this database: a) CMATERdb 3.1.1 for Bangla numerals, b) CMATERdb 3.2.1 for Devanagari (Hindi) numerals, c) CMATERdb 3.3.1 for Arabic numerals, d) CMATERdb 3.4.1 for Telugu numerals and e) CMATERdb 3.1.2 for Bangla characters. Sample images from the datasets are shown in Fig. 6(a-e).

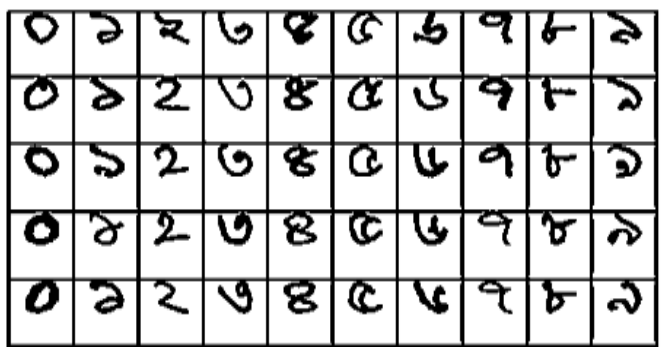
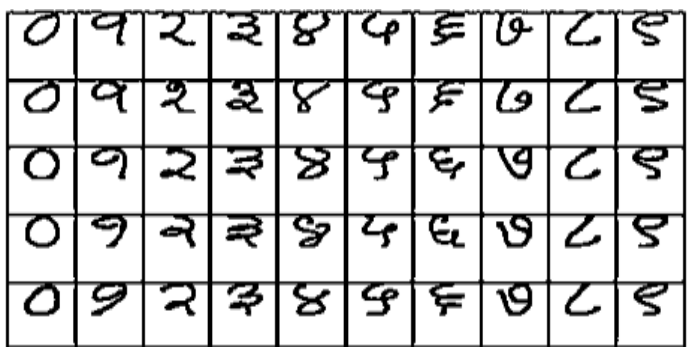
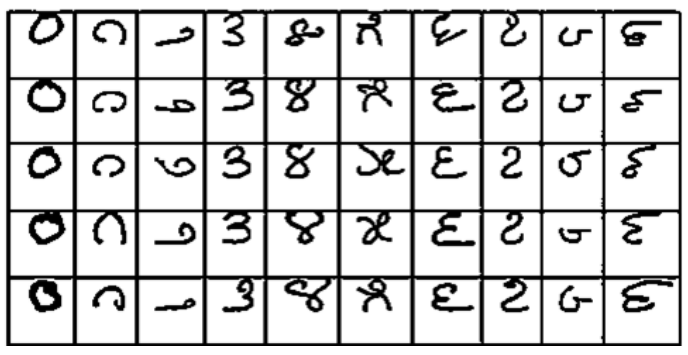

(a) (b) (c)

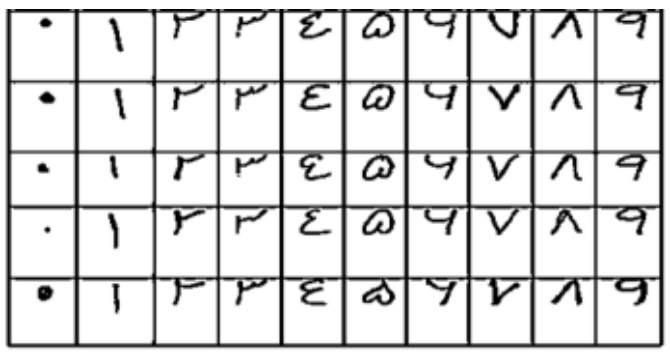
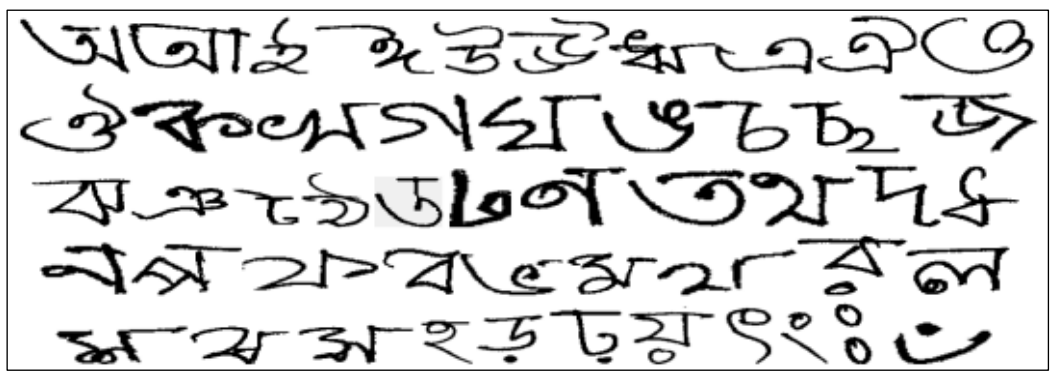

(d) (e)

**Fig. 6.** Sample images from CMATERdb3 database: (a) Bangla numeral, (b) Devanagari numeral, (c) Telugu numeral, (d) Arabic numeral, and (e) Bangla characters

TABLE I

DATABASE DETAILS FOR EXPERIMENTATION

| Database | Number of Classes | Training Size | Test Size | Total |
|---|---|---|---|---|
| CMATERdb 3.1.1 Bangla numeral | 10 | 4000 | 2000 | 6000 |
| CMATERdb 3.2.1 Devanagari numeral | 10 | 2000 | 1000 | 3000 |
| CMATERdb 3.3.1 Arabic numeral | 10 | 2000 | 1000 | 3000 |
| CMATERdb 3.4.1 Telugu numeral | 10 | 2000 | 1000 | 3000 |
| CMATERdb 3.1.2 Bangla character | 50 | 12000 | 3000 | 15000 |
| ISI Kolkata Bangla numeral | 10 | 19392 | 4000 | 23392 |
| ISI Kolkata Odia numeral | 10 | 4000 | 1000 | 5000 |
| IITBBS Odia numeral | 10 | 4000 | 1000 | 5000 |
| IITBBS Odia character | 68 | 8160 | 2040 | 10200 |
| MNIST English numeral | 10 | 60000 | 10000 | 70000 |

### *3.2.2. ISI Kolkata database*

There exists two numeral datasets, one for Bangla and the other for Odia language. The ISI Kolkata Bangla handwritten numeral database [26] is created from 1106 individuals and few samples are shown in Fig. 7(a). The database includes images from 465 mail pieces, and 268 job application forms. Similarly, the ISI Kolkata Odia handwritten numeral database [50] consists of 5000 images with 500 images per class (Fig. 7(b)). This database is created by collecting numerals from more than 105 postal mail documents and 166 job application forms. All the documents are scanned using a flatbed scanner at 300dpi resolution.

### 3.2.3. *IITBBS database*

Odia is one of the six classical languages of India. The handwritten images in IITBBS Odia numeral [23] and character [51] database are collected from more than 500 individuals. The images contain significant amount of noise and deformation due to unconstrained handwriting variations. The images are digitised with 300 and 600dpi resolution. A sample of handwritings from the numeral database is shown in Fig. 8(a). There are 68 character classes in the database comprising basic (vowels and consonants) as well as modified and compound Odia handwritten characters. With 150 patterns in each class, the database size is $150 \times 68 = 10200$. A sample of images from the character database is shown in Fig. 8(b).

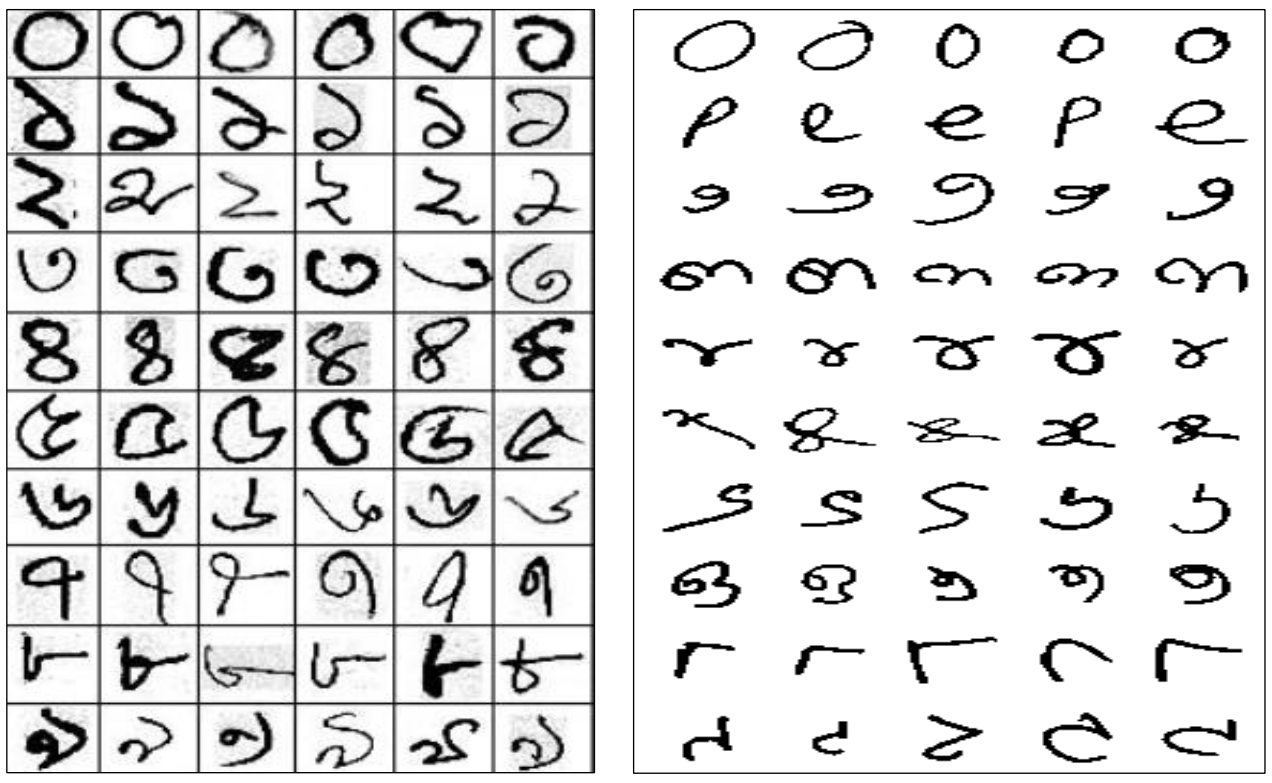

(a) (b)

**Fig. 7.** Sample handwritten images from ISI Kolkata database: (a) Bangla numerals, and (b) Odia numerals

### 3.2.4. *MNIST database*

The MNIST database [52] comprises of 60,000 samples of handwritten numerals collected from approximately 250 individuals for training. The test set consists of 10,000 images. The sets of writers for training and test set are disjoint. The original images of this database are centered by calculating and translating the center of mass of the pixels in a grid size of $28 \times 28$ pixels. A sample of MNIST database images are shown in Fig. 9.

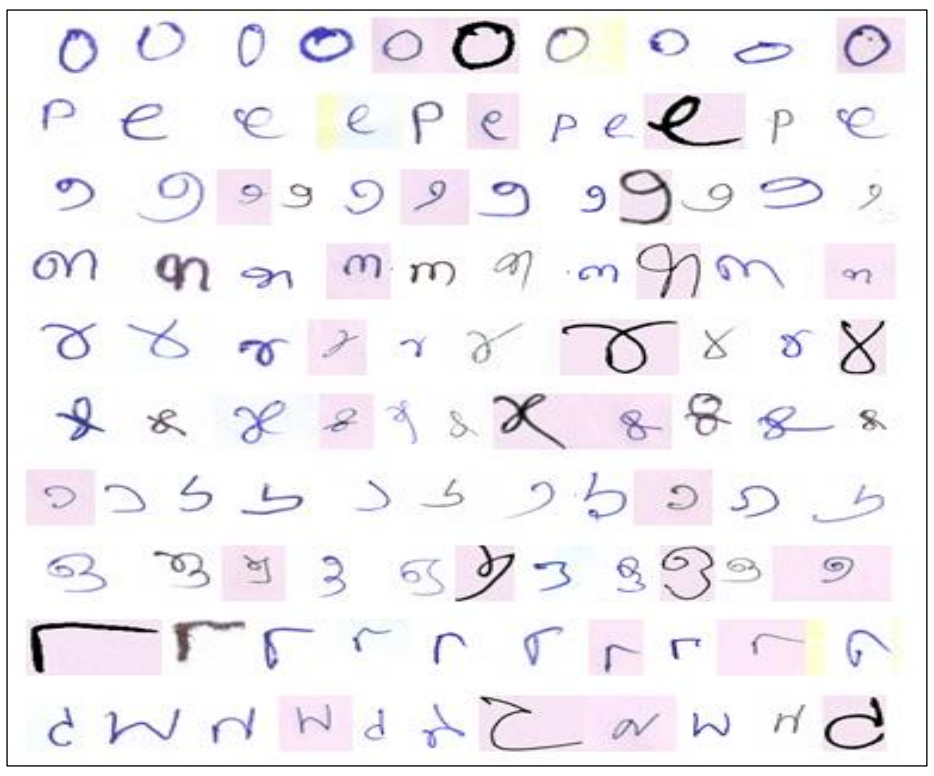

(a)

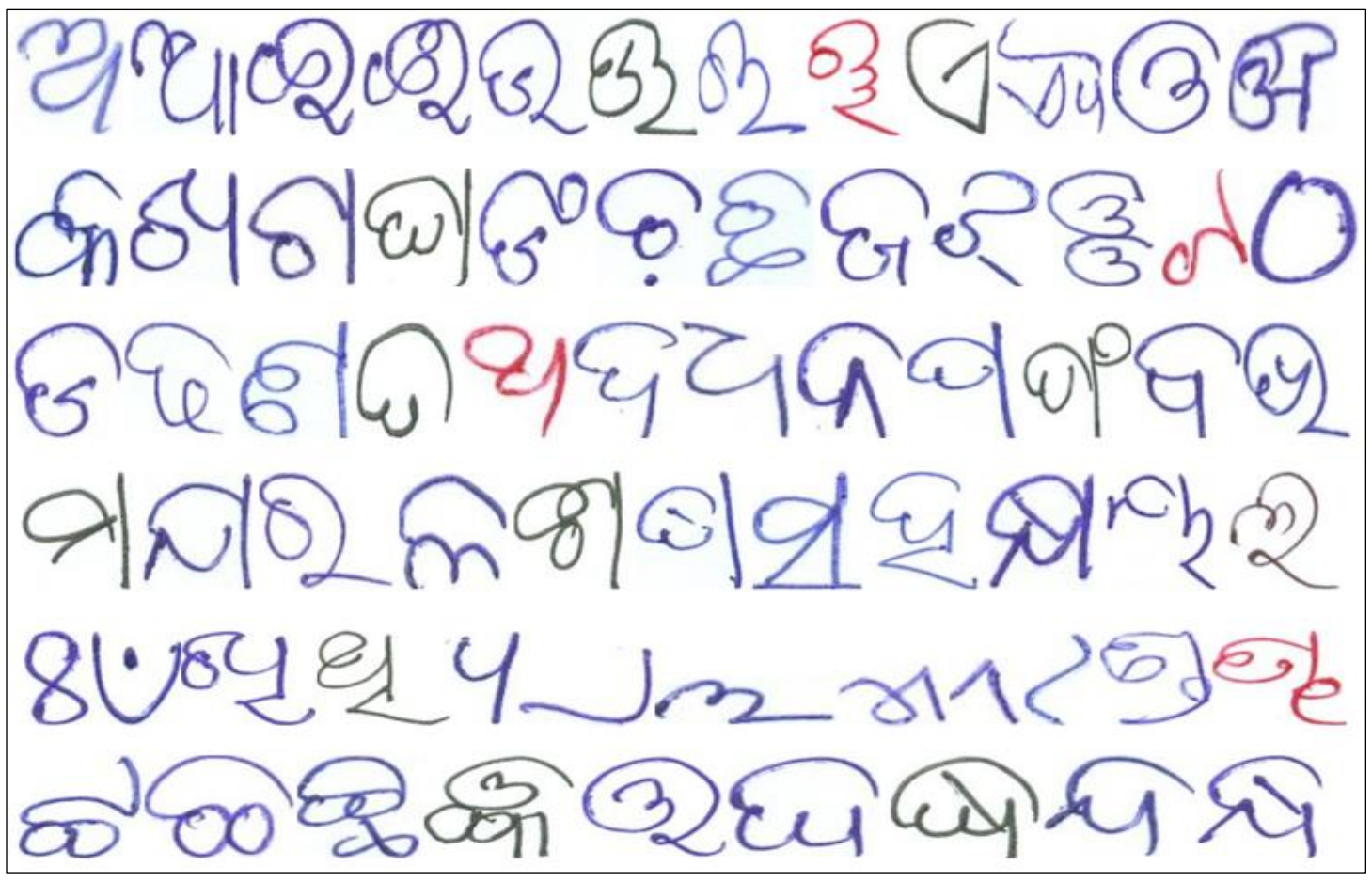

(b)

**Fig. 8.** Sample handwritten images from (a) IITBBS Odia numeral, (b) IITBBS Odia character

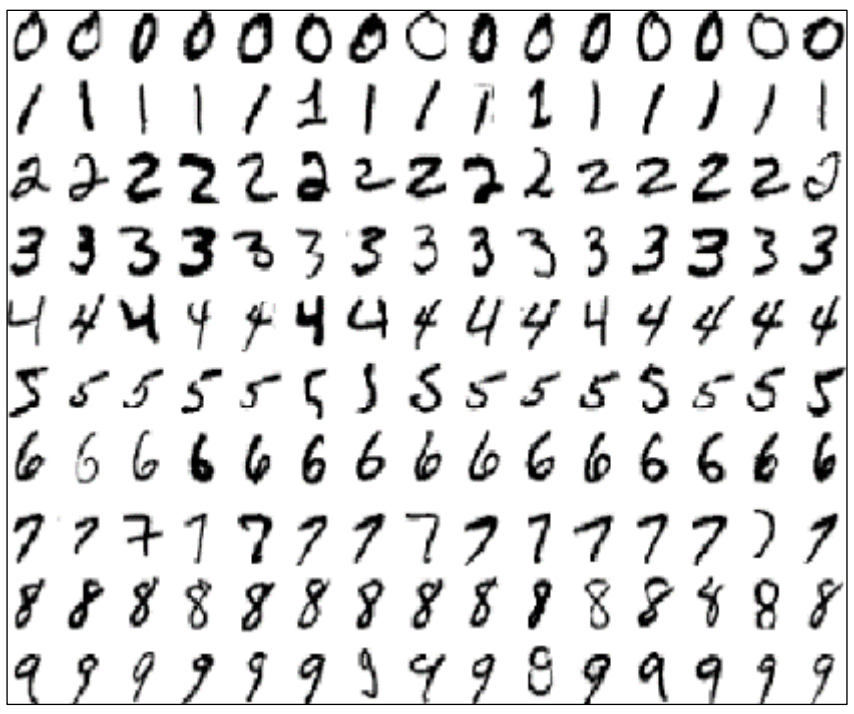

**Fig. 9**. Sample handwritten images from MNIST database

### *3.3. Recognition Performance*

A 5-fold cross validation technique is adopted by taking different subsets from each database as training and test sets for verifying robustness of the system. It is to be noted that the total number of training samples and test samples are kept the same as described for respective databases in order to have uniformity in performance comparison. The performance of our proposed recognition system on all the benchmark databases is calculated using three figure of merits: True Positive (*TP*), False Positive (*FP*) and False Negative (*FN*). For a $N$ class recognition system, the confusion matrix can be denoted as $C_{N\times N}:\left\{c_{ij}\right\}, i=1,...,N,\, j=1,...,N$ where $c_{ij}$ refers to the number of patterns belonging to class $i$ and recognized as class $j$. For any class $k$, the three figure of merit metrics can be defined as:

$$TP_k = c_{kk}, \quad FP_k = \sum_{i=1, i\neq k}^{i=N} c_{ik}, \quad FN_k = \sum_{j=1, j\neq k}^{j=N} c_{kj} \tag{21}$$

Two standard parameters, Recall and Precision, are also computed for establishing measurable performance of the proposed recognizer:

$$Recall_k = \frac{TP_k}{TP_k + FN_k}$$

$$Precision_k = \frac{TP_k}{TP_k + FP_k} \tag{22}$$

Finally, the classification accuracy is calculated as

$$Accuracy = \left( \frac{1}{N} \sum_{k=1}^{N} \frac{TP_k}{TP_k + FN_k} \right) \times 100 \quad . \tag{23}$$

After 5-fold cross validation, the average accuracy is found out which is the final classification rate for a particular recognition system. The highest classification accuracies using the proposed feature representation and different classifier models on all the databases are given in Table X.

TABLE X

RECOGNITION ACCURACY USING THE PROPOSED METHOD: 5-FOLD CROSS-VALIDATION ON ALL DATABASES

| **Database** | **Folds** | | | | | **Average** |
|---|---|---|---|---|---|---|
| | **1** | **2** | **3** | **4** | **5** | |
| CMATERdb 3.1.1 Bangla numeral | 97.45 | 98.8 | 98.55 | 98.35 | 99.05 | 98.44 |
| CMATERdb 3.2.1 Hindi numeral | 98.8 | 99.1 | 98.5 | 99.1 | 99.3 | 98.96 |
| CMATERdb 3.3.1 Arabic numeral | 98.7 | 99.2 | 98.9 | 98.4 | 97.9 | 98.62 |
| CMATERdb 3.4.1 Telugu numeral | 99.3 | 99 | 98.9 | 99.1 | 98.7 | 99.00 |
| CMATERdb 3.1.2 Bangla character | 95.97 | 93.77 | 94.5 | 93.97 | 95.7 | 94.78 |
| ISI Kolkata Bangla numeral | 99.475 | 99.2 | 99.275 | 99.7 | 99.25 | 99.38 |
| ISI Kolkata Odia numeral | 98.6 | 99.5 | 99.5 | 99.3 | 99.2 | 99.22 |
| IITBBS Odia numeral | 98.3 | 99.1 | 98.9 | 98.6 | 98.7 | 98.72 |
| IITBBS Odia character | 93.873 | 94.46 | 94.069 | 92.108 | 91.667 | 93.24 |
| MNIST | 99.37 | 99.48 | 99.26 | 99.39 | 99.48 | 99.40 |

### 3.4. *Performance Comparison*

Four standard classification models along with the proposed feature extraction are used for experimental validation: nearest neighborhood classifier, support vector machine (SVM), random forest (RF), and modified quadratic discriminant function (MQDF) classifier [57]. The experimental results are compared with the state-of-the-art character recognition approaches existing for the databases.

TABLE XI

PERFORMANCE COMPARISON ON CMATERDB DATABASES

| **Database** | **Method** | **Features** | **Classifier** | **Accuracy (%)** |
|---|---|---|---|---|
| CMATERdb 3.1.1 Bangla numeral | Basu et al. [11] | QTLR | MLP | 96.45 |
| | Das et al. [53] | SOLR | SVM | 97.70 |
| | Das et al. [31] | MPCA + QTLR | SVM | **98.55** |
| | Proposed Method | Tetrolet + SCC | MQDF | 97.16 |
| | | | Random Forest | 97.63 |
| | | | SVM | 98.20 |
| | | | Nearest Neighbor | **98.44** |
| CMATERdb 3.2.1 Devanagari numeral | Das et al. [31] | MPCA + QTLR | SVM | **98.70** |
| | Proposed Method | Tetrolet + SCC | MQDF | 96.54 |
| | | | SVM | 97.15 |
| | | | Random Forest | 97.38 |
| | | | Nearest Neighbor | **98.96** |
| CMATERdb 3.3.1 Arabic numeral | Das et al. [15] | Shadow, Octant centroids | MLP | 95.80 |
| | Das et al. [31] | MPCA + QTLR | SVM | **98.40** |
| | Proposed Method | Tetrolet + SCC | MQDF | 96.80 |
| | | | Random Forest | 97.71 |
| | | | SVM | 98.47 |
| | | | Nearest Neighbor | **98.62** |
| CMATERdb 3.4.1 Telugu numeral | Das et al. [31] | MPCA + QTLR | SVM | **99.10** |
| | Proposed Method | Tetrolet + SCC | MQDF | 98.25 |
| | | | Random Forest | 98.46 |
| | | | Nearest Neighbor | 98.92 |
| | | | SVM | **99.00** |
| CMATERdb 3.1.2 Bangla character | Basu et al. [54] | Shadow, centroid, longest-run | MLP | 75.05 |
| | Rahman et al. [55] | Matra based | MLP | **88.38** |
| | Proposed Method | Tetrolet + SCC | MQDF | 88.33 |
| | | | Random Forest | 90.58 |
| | | | SVM | 92.15 |
| | | | Nearest Neighbor | **94.78** |

Table XI compares the classification accuracy using the proposed recognition system on five CMATERdb databases with the existing recognition results. The highest accuracy was achieved by nearest neighbor classifier on CMATERdb 3.1.1 (98.44%), CMATERdb 3.2.1 (98.96%), CMATERdb 3.3.1 (98.62%) and CMATERdb 3.1.2 (94.78%) databases. Similarly, the highest accuracy reported on CMATERdb 3.4.1 Telugu numeral database (99.00%) was by using the SVM classifier. From Table XI, it is evident that the proposed method outperforms the state-of-the-art recognition accuracies on Devanagari numeral, Arabic numeral and Bangla character datasets of CMATER database. The classification accuracies are comparable to the existing results in cases of Bangla and Telugu handwritten numerals of the same database.

TABLE XII

PERFORMANCE COMPARISON ON ISI KOLKATA DATABASES

| Database | Method | Features | Classifier | Accuracy (%) |
|---|---|---|---|---|
| ISI Kolkata Bangla numeral | Pal et al. [56] | Water reservoir | Binary tree | 92.80 |
| | Wen et al. [16] | Statistical | KBD-G | 96.91 |
| | Roy et al. [12] | Directional | MLP | 96.93 |
| | Basu et al. [11] | QTLR | SVM | 97.15 |
| | Purkait and Chanda [17] | Morphological | MLP | 97.75 |
| | Bhattacharya and Chaudhuri [26] | Wavelet | MLP | 98.20 |
| | Liu and Suen [13] | Gradient | DLQDF | **99.40** |
| | Proposed Method | Tetrolet + SCC | Random Forest | 96.76 |
| | | | MQDF | 97.20 |
| | | | Nearest Neighbor | 98.65 |
| | | | SVM | **99.38** |
| ISI Kolkata Odia numeral | Bhowmik et al. [21] | Scalar | HMM | 90.50 |
| | Dash et al. [24] | Gestalt | Complexity Metric | 92.82 |
| | Roy et al. [22] | Directional | QDF | 94.81 |
| | Pal et al. [20] | Directional | MQDF | 98.40 |
| | Dash et al. [14] | Hybrid | DLQDF | 98.50 |
| | Dash et al. [23] | AZNRST | Nearest Neighbor | **99.10** |
| | Proposed Method | Tetrolet + SCC | Random Forest | 97.60 |
| | | | SVM | 98.84 |
| | | | MQDF | 98.95 |
| | | | Nearest Neighbor | **99.22** |

TABLE XIII

PERFORMANCE COMPARISON ON IITBBS DATABASES

| Database | Method | Features | Classifier | Accuracy (%) |
|---|---|---|---|---|
| IITBBS Odia character | Pal et al. [10] | Curvature | QDF | **91.38** |
| | Proposed Method | Tetrolet + SCC | SVM | 89.75 |
| | | | Random Forest | 90.06 |
| | | | MQDF | 91.18 |
| | | | Nearest Neighbor | **93.24** |
| IITBBS Odia numeral | Bhowmik et al. [21] | Scalar | HMM | 88.74 |
| | Dash et al. [24] | Gestalt | Complexity Metric | 91.40 |
| | Roy et al. [22] | Directional | QDF | 94.12 |
| | Pal et al. [20] | Directional | MQDF | 96.30 |
| | Dash et al. [14] | Hybrid | DLQDF | 98.28 |
| | Dash et al. [23] | AZNRST | Nearest Neighbor | **98.60** |
| | Proposed Method | Tetrolet + SCC | Random Forest | 95.79 |
| | | | MQDF | 97.14 |
| | | | SVM | 98.36 |
| | | | Nearest Neighbor | **98.72** |

TABLE XIV

PERFORMANCE COMPARISON ON MNIST DATABASE

| Database | Method | Features | Classifier | Accuracy (%) |
|---|---|---|---|---|
| MNIST | Cecotti [28] | Active graphs | Nearest Neighbor | 99.19 |
| | Belongie et al. [29] | Shape Context | Nearest Neighbor | 99.37 |
| | Liu et al. [30] | Directional | SVC | 99.39 |
| | Ciresan et al. [18] | Raw image | MCDNN | **99.77** |
| | Proposed Method | Tetrolet + SCC | MQDF | 98.98 |
| | | | Nearest Neighbor | 99.23 |
| | | | SVM | 99.27 |
| | | | Random Forest | **99.40** |

Similarly, Table XII summarizes the classification accuracies on two datasets of ISI Kolkata database. A SVM classifier achieves 99.38% accuracy which is at par with the highest accuracy reported on handwritten Bangla numerals. A nearest neighbor classifier outperforms (99.22%) the previous highest accuracy (99.10%) on ISI Kolkata Odia numeral database.

From Table XIII, it is clear that the proposed OCR system performs better than existing techniques on IITBBS Odia numeral database, achieving highest accuracy using nearest neighbor classifier (98.72%). As IITBBS Odia character database is a new open access database, there are no existing reports that have validated their methods on the database. We implemented the method designed for Odia characters in [10] and an accuracy of 93.24% is achieved on the new IITBBS Odia character database. On the other hand, the random forest classifier achieves highest recognition rate in case of MNIST database (99.40%) which is very close to the state-of-the-art recognition result (Table XIV).

Moreover, we compare the classification accuracy of the proposed feature representation with other sparse based techniques such as PCA, SparsePCA [37] and SparseLDA [38] for different values of sparsity parameter $k$ . The comparison results summarized in Fig. 10 indicate the superiority of the new method in four different databases over four different scripts. In addition, to categorically show the superiority of sparse coded Tetrolet feature extraction technique, we compared the recognition result on all the databases using other spatio-spectral feature representations on handwritten character recognition. The multiresolution techniques such as Wavelets, Stockwell transform and Slantlets are implemented for performance comparison. The comparison histograms are given in Fig. 11.

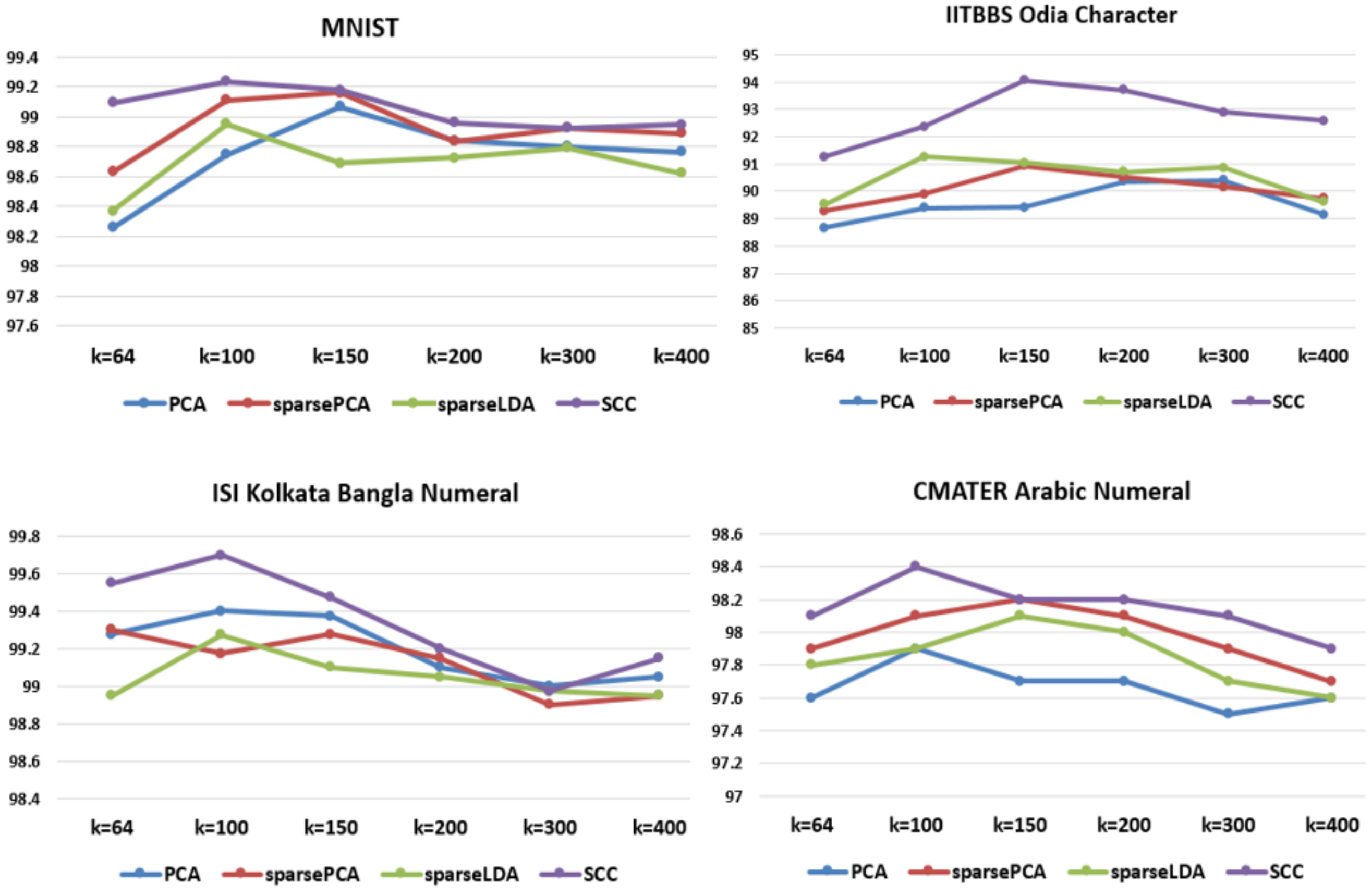

**Fig. 10.** Comparison of sparse based coding techniques on different databases proving superiority of SCC

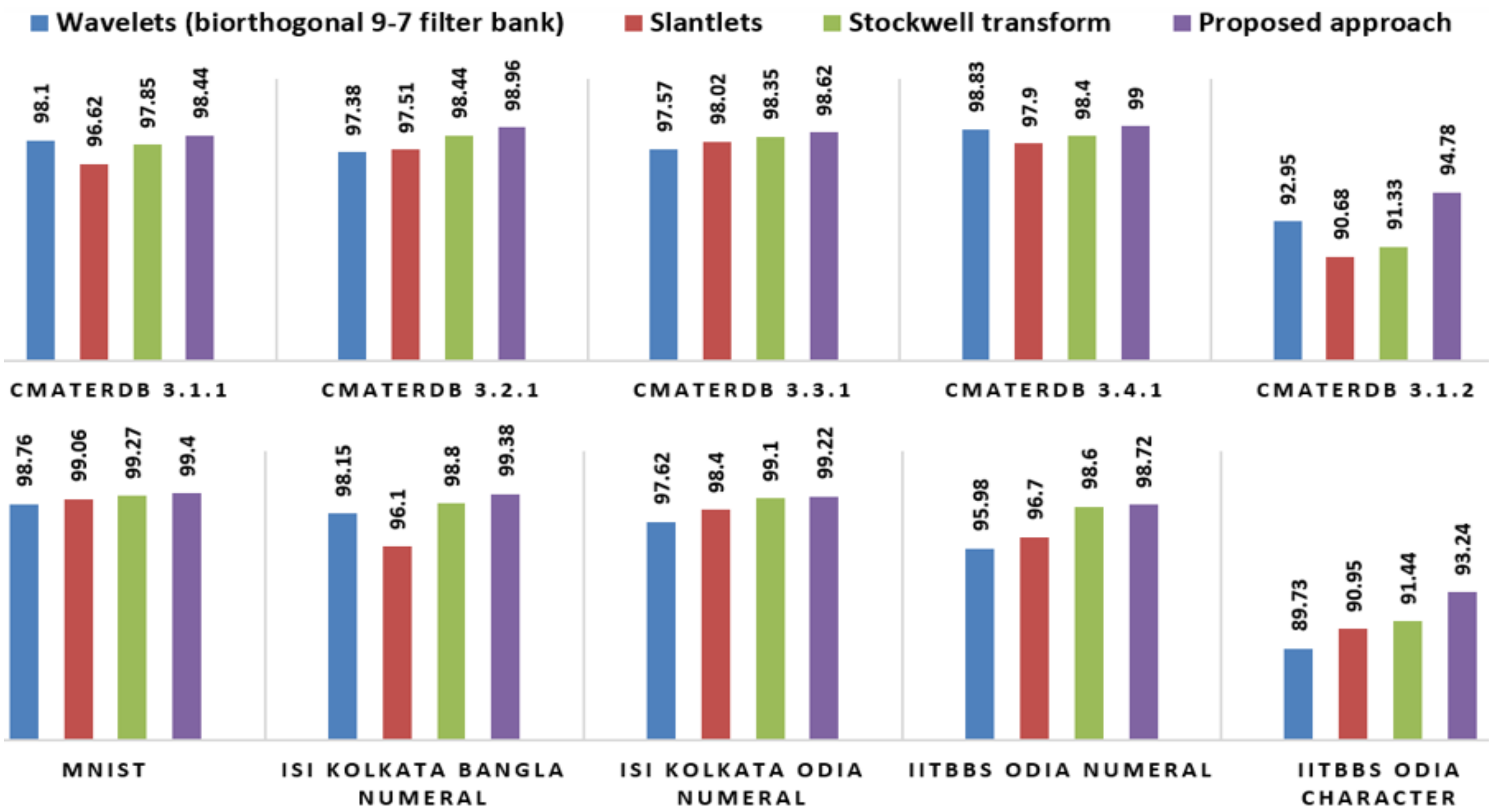

**Fig. 11.** Histograms representing comparison of classification accuracies using Daubechies wavelet, Slantlet, Stockwell and the proposed Tetrolet feature representation

## 4. Conclusion

In this paper, a new approach to unconstrained handwritten character recognition is proposed combining sparse concept coding and Tetrolet transform based space-frequency decomposition. The exhaustive experimentation on ten standard databases spanning six different scripts provides the basis for performance validation and benchmarking. The proposed recognition system is very fast in recognizing test character images (33ms on an average to classify a handwritten test image); thus can be designed for a real-time recognition task. The accuracies obtained are state-of-the-art and can further be extended to other languages. Evaluation and database creation on more scripts can be taken up in future so that the proposed system can evolve to achieve human like accuracy which still remains a challenging task.